\documentclass[runningheads]{llncs}
\usepackage{graphicx}
\usepackage{comment}
\usepackage{amsmath,amssymb} 
\usepackage{color}
\usepackage{subfigure}
\usepackage{etoolbox}
\usepackage{booktabs}
\usepackage{multirow}
\usepackage{cite}
\usepackage{url}

\begin{document}
	

\newcommand{\MyMapTemplatePrefix}[4]{\expandafter#1\csname#3#4\endcsname{#2{#4}}}
\newcommand{\MyMapTemplatePrefixNew}[5]{\expandafter#1\csname#4#5\endcsname{#2{#3{#5}}}}
\forcsvlist{\MyMapTemplatePrefix {\def} {\mathbf} {}} {A,B,C,D,E,F,G,H,I,J,K,L,M,N,O,P,Q,R,S,T,U,V,W,X,Y,Z} 
\forcsvlist{\MyMapTemplatePrefix {\def} {\mathbf} {}} {a,b,c,d,e,f,g,h,i,j,k,l,m,n,o,p,q,r,s,t,u,v,w,x,y,z,1,0}
\forcsvlist{\MyMapTemplatePrefix {\def} {\widetilde} {wt}} {A,B,C,D,E,F,G,H,I,J,K,L,M,N,O,P,Q,R,S,T,U,V,W,X,Y,Z}
\forcsvlist{\MyMapTemplatePrefix {\def} {\widetilde} {wt}} {a,b,c,d,e,f,g,h,i,j,k,l,m,n,o,p,q,r,s,t,u,v,w,x,y,z} 
\forcsvlist{\MyMapTemplatePrefixNew {\def} {\widetilde}{\mathbf} {tb}} {A,B,C,D,E,F,G,H,I,J,K,L,M,N,O,P,Q,R,S,T,U,V,W,X,Y,Z}
\forcsvlist{\MyMapTemplatePrefixNew {\def} {\widetilde}{\mathbf} {tb}} {a,b,c,d,e,f,g,h,i,j,k,l,m,n,o,p,q,r,s,t,u,v,w,x,y,z}
\forcsvlist{\MyMapTemplatePrefix {\def} {\widehat} {wh}} {A,B,C,D,E,F,G,H,I,J,K,L,M,N,O,P,Q,R,S,T,U,V,W,X,Y,Z}
\forcsvlist{\MyMapTemplatePrefix {\def} {\widehat} {wh}} {a,b,c,d,e,f,g,h,i,j,k,l,m,n,o,p,q,r,s,t,u,v,w,x,y,z}
\forcsvlist{\MyMapTemplatePrefixNew {\def} {\widehat}{\mathbf} {hb}} {A,B,C,D,E,F,G,H,I,J,K,L,M,N,O,P,Q,R,S,T,U,V,W,X,Y,Z}
\forcsvlist{\MyMapTemplatePrefixNew {\def} {\widehat}{\mathbf} {hb}} {a,b,c,d,e,f,g,h,i,j,k,l,m,n,o,p,q,r,s,t,u,v,w,x,y,z}
\forcsvlist{\MyMapTemplatePrefix {\def} {\mathcal}{mc}} {A,B,C,D,E,F,G,H,I,J,K,L,M,N,O,P,Q,R,S,T,U,V,W,X,Y,Z}
\forcsvlist{\MyMapTemplatePrefix {\def} {\mathcal}{mc}} {a,b,c,d,e,f,G,H,I,J,K,L,M,N,O,P,Q,R,s,T,U,V,W,X,Y,Z}
\forcsvlist{\MyMapTemplatePrefix {\def} {\mathbb} {mb}} {A,B,C,D,E,F,G,H,I,J,K,L,M,N,O,P,Q,R,S,T,U,V,W,X,Y,Z}
\def\tp{^\top}
\def\st{\text{s.t.~}}
\def\bmu{\boldsymbol{\mu}} \def\bSigma{\boldsymbol{\Sigma}}
\def\ie{\emph{i.e.}}
\def\etal{\emph{et al.}}
\def\ptl{\partial~}
\def\eg{\emph{e.g.}}
\def\phitb{\widetilde{\phi}}
\def\bphi{\boldsymbol{\phi}}
\def\bPhi{\boldsymbol{\Phi}}
\def\tlam{\tilde{\lambda}}
\def\blam{\boldsymbol{\lambda}}
\newcommand{\fix}{\marginpar{FIX}}
\newcommand{\new}{\marginpar{NEW}}
\newtheorem{thm}{Theorem}
\newtheorem{defn}[thm]{Definition}
\newtheorem{prop}{Proposition}
\newcommand{\tabincell}[2]{\begin{tabular}{@{}#1@{}}#2\end{tabular}}
\newcommand{\tabnote}[1]{\par\medskip\parbox{\textwidth}{#1}}
\hyphenation{all resp images image matrix sparse SFRD STFRD PMML its appearance transform}

\pagestyle{headings}
\mainmatter
\def\ECCVSubNumber{5066}  

\title{Graph Wasserstein Correlation Analysis for Movie Retrieval} 

\titlerunning{Graph Wasserstein Correlation Analysis for Movie Retrieval}
%
\author{Xueya Zhang \and
Tong Zhang\and
Xiaobin Hong\and
Zhen Cui\and Jian Yang}
\authorrunning{Xueya Zhang et al.}
%
\institute{Key Lab of Intelligent Perception and Systems for High-Dimensional Information of Ministry of Education, School of Computer Science and Engineering, Nanjing University of Science and Technology\\
\email{\{zhangxueya,tong.zhang,xbhong,zhen.cui,csjyang\}@njust.edu.cn}}
\maketitle

\newcommand\blfootnote[1]{%
	\begingroup 
	\renewcommand\thefootnote{}\footnote{#1}%
	\addtocounter{footnote}{-1}%
	\endgroup 
}
\blfootnote{Xueya Zhang and Tong Zhang have equal contributions.}
\blfootnote{Corresponding author: zhen.cui@njust.edu.cn.}
\begin{abstract}
Movie graphs play an important role to bridge heterogenous modalities of videos and texts in human-centric retrieval. In this work, we propose Graph Wasserstein Correlation Analysis (GWCA) to deal with the core issue therein, i.e, cross heterogeneous graph comparison. Spectral graph filtering is introduced to encode graph signals, which are then embedded as probability distributions in a Wasserstein space, called graph Wasserstein metric learning. Such a seamless integration of graph signal filtering together with metric learning results in a surprise consistency on both learning processes, in which the goal of metric learning is just to optimize signal filters or vice versa. Further, we
derive the solution of the graph comparison model as a classic generalized eigenvalue decomposition problem, which has an exactly closed-form solution. Finally, GWCA together with movie/text graphs generation are unified into the framework of movie retrieval to evaluate our proposed method. Extensive experiments on MovieGrpahs dataset demonstrate the effectiveness of our GWCA as well as the entire framework.
\keywords{graph Wasserstein metric \and graph correlation analysis \and movie retrieval}
\end{abstract}

\section{Introduction}

Nowadays, people show growing enthusiasm in searching desired movie clips, which contain either attractive plots or funny dialogue with vivid performance of actors, for multiple purposes including materials accumulation for presentation and entertainment.  However, in many cases, they can just describe their understanding/impression of plots or dialogue content of those target clips, but are hardly accessible to the exact movie names or frame locations. This makes it time/energy-consuming and tedious to search the desired clips by manually browsing those movies one by one. Consequently, automatic movie-text retrieval become quite necessary and meaningful.

Among movie retrieval, the elements mainly consist of visual videos and descriptive texts, which have been investigated in some cross tasks such as video description{~\cite{chen2014learning}} and video/image query and answer (Q \& A){~\cite{tapaswi2016movieqa}}. Most methods take some sophisticated dynamic models, e.g.,gated recurrent unit (GRU){~\cite{chung2014empirical}} and long-short term memory (LSTM){~\cite{xingjian2015convolutional}}, to capture the dynamics within both videos and texts, and then bridge them based on those obtained representation. However, these do not cater to flexible movie contour search, where some actors might be only posed by one searcher. Just to address this case, recently MovieGraphs dataset{~\cite{vicol2018moviegraphs}} is successfully  initiated with annotated graphs to describe the interactions of entities in movie clips, and provides rather appropriate evaluations on more flexible movie-description retrieval for boosting machine understanding on movie clips.

Motivated by this case, in this work, we follow the technique line of graph modeling, which is more versatile to describe structured information in human-centric situation of movie graph retrieval. As a universal tool, graph can represent various data in the real
world by defining nodes and edges that reveal multiple relationships between objects. For one given movie clip, those actors or other entities could be understood as nodes, their interactions may be defined as the edge connections. Accordingly, the text description can also be modeled with graph structure. Hence, the task of movie retrieval can be converted into the problem of graph searching, whose core issue is the  comparisons between graph structured data. The inter-graph comparison contains two crucial problems: graph signal processing and graph distance metric. The former focuses on how to mine useful information from graph structure data, while the latter concerns the measurement of two graphs. On one hand, the obstacle to encode graph signals is not only to process graph signals as discrete time signal but also need model dependencies arising from irregular data. On the other hand, for graph structured data, Euclidean metrics fundamentally limit the ability to capture latent semantic structures, which however need not conform to Euclidean spatial assumptions. Further, could graph signal processing be seamlessly integrated with graph distance metric learning for more effective comparisons between graph data?

In this paper, we propose a Graph Wasserstein Correlation Analysis (GWCA) method to deal with the comparisons of pairwise movie graphs. The proposed GWCA elegantly formulates graph signal encoding together with graph distance metric learning into a unified model. Inspired by the recent spectral graph theory, we encode graph structure data with spectral graph filtering, which generalizes the previous classic signal processing. Instead of direct frequency domain, we take an approximation strategy, i.e., the polynomial of graph Laplacian, to efficiently encode graph data. The encoded signals of graph are embedded as probability distributions in a Wasserstein space, which is much larger and more flexible than Euclidean space. Accordingly, the distance between graph data is defined in Wasserstein space, which is called Wasserstein metric. Such a metric can not only captures the similarity of the distributions of graph signals, but also be able to preserve the transitivity in embedding space. In this way, graph signal filtering and Wasserstein metric learning are jointly encapsulated into a unified mathematic model, which efficiently preserves the first-order and second-order proximity of the nodes of graph, empowering the learned node representations to reflect both graph topology structure. Surprisingly, we derive this model as a classic eigenvalue decomposition problem with closed-form solution, where the solution is just associated with graph encoding. Finally, our GWCA is used to movie graph retrieval, where multiple heterogeneous graphs are built and crossly-compared, e.g., annotation graph versus description graph, video graph versus annotation graph, etc. Extensive experiments on MovieGraphs dataset demonstrate the effectiveness of our proposed method, and new state-of-the-art results are also achieved.

In summary, our contribution are three folds:
\begin{itemize}
	\item We propose an elegant inter-graph comparison model by seamlessly integrating graph signals filtering together with graph Wasserstein metric learning, where the latter is just the optimization of the former.
	\item We derive the solution of model as a classic generalized eigenvalue decomposition problem, which has an exact closed-form solution.
	\item We design an entire framework for movie retrieval including graph generation and GWCA, and finally validate the effectiveness of our proposed method.
\end{itemize}

\section{Related Work}
Most relevant works are proposed to inference across vision and text, where multiple tasks are tackled including image-text modeling{~\cite{farhadi2010every,ordonez2011im2text,yang2011corpus}}, video/image query and answer (Q \& A){~\cite{tapaswi2016movieqa}} and video-text retrieval~\cite{sankar2009subtitle,cour2008movie}. For image-text understanding, a majority of work generate descriptive sentences for vision, and especially, {~\cite{chen2014learning}} including sentence generation and image retrieval to find the bi-directional mapping between images and their textual descriptions. For video based works,{~\cite{bojanowski2013finding}} focus on understanding action of characters with scripts and {~\cite{ding2010learning}} learn the relations among actors. {~\cite{laptev2008learning}} proposed the method using the retrieved action samples for visual learning and achieving action classification based on texts. In~\cite{pan2016jointly}, authors propose an LSTM with visual semantic embedding method. Recently, Vicol \etal~\cite{vicol2018moviegraphs} proposed a new dataset MovieGraphs for retrieving videos and text with graphs, which also shows graphs containing sufficient information help us to understand the video and text better.

\textbf{Graph Signal Processing.} Graphs are generic data representation forms, which describe the geometric structures of data domains effectively. From the perspective of graph signal processing, the data on these graphs can be regarded as a finite collection of samples, and the sample at each vertex in the graph is graph signal.{~\cite{shuman2013emerging}} concluded that spectral graph theory is regarded as the tool for defining the frequency spectra , and as an extension of the Fourier transform of the graph. It benefits the construction of expander graphs{~\cite{hoory2006expander}}, spectral clustering{~\cite{von2007tutorial}} and so on, including definitions and notations such as the Non-Normalized Graph Laplacian and Graph Fourier Transform. References {~\cite{tenenbaum2000global, roweis2000nonlinear, belkin2003laplacian,donoho2003hessian}} generate low dimensional representations for  high-dimensional data through spectral graph theory and the graph Laplacian {\cite{chung1996lectures}}, projecting the data on a low-dimensional subspace generated by a small subset of the Laplacian eigenbasis{~\cite{belkin2003laplacian}}.\\
Generalized operators like filtering and translation then become the basis of developing the localized, multi-scale transforms. In{~\cite{buades2005review}}, the basic graph spectral filtering enable discrete versions of continuous filtering, known as Gaussian smoothing, bilateral filtering, anisotropic diffusion, and non-local means filtering. Especially, Bruna \etal {~\cite{Bruna2014Spectral}} consider possible generalizations of CNNs, which extends convolution networks to graph domains. Then Defferrard \etal {~\cite{defferrard2016convolutional}} proposed a fast spectral filter, which use the Chebyshev polynomial approximation so that they are of the same linear computation complexity. Kipf \etal{~\cite{kipf2016semi}} motivate the convolutional architecture with a localized first-order approximation of spectral graph convolutions. In particular, {~\cite{lafon2006diffusion,ron2011relaxation,narang2009lifting}} propose the literature of graph coarsening, downsampling and reduction. These graph modeling methods have also been applied to many tasks, such as node classification{~\cite{zhao2019hashing,jiang2019gaussian}}, action recognition{~\cite{li2019spatio}} and user recommendation{~\cite{zhang2020cross}}.

\begin{figure*}[!t]
	\centering
	\includegraphics[width=0.5\linewidth]{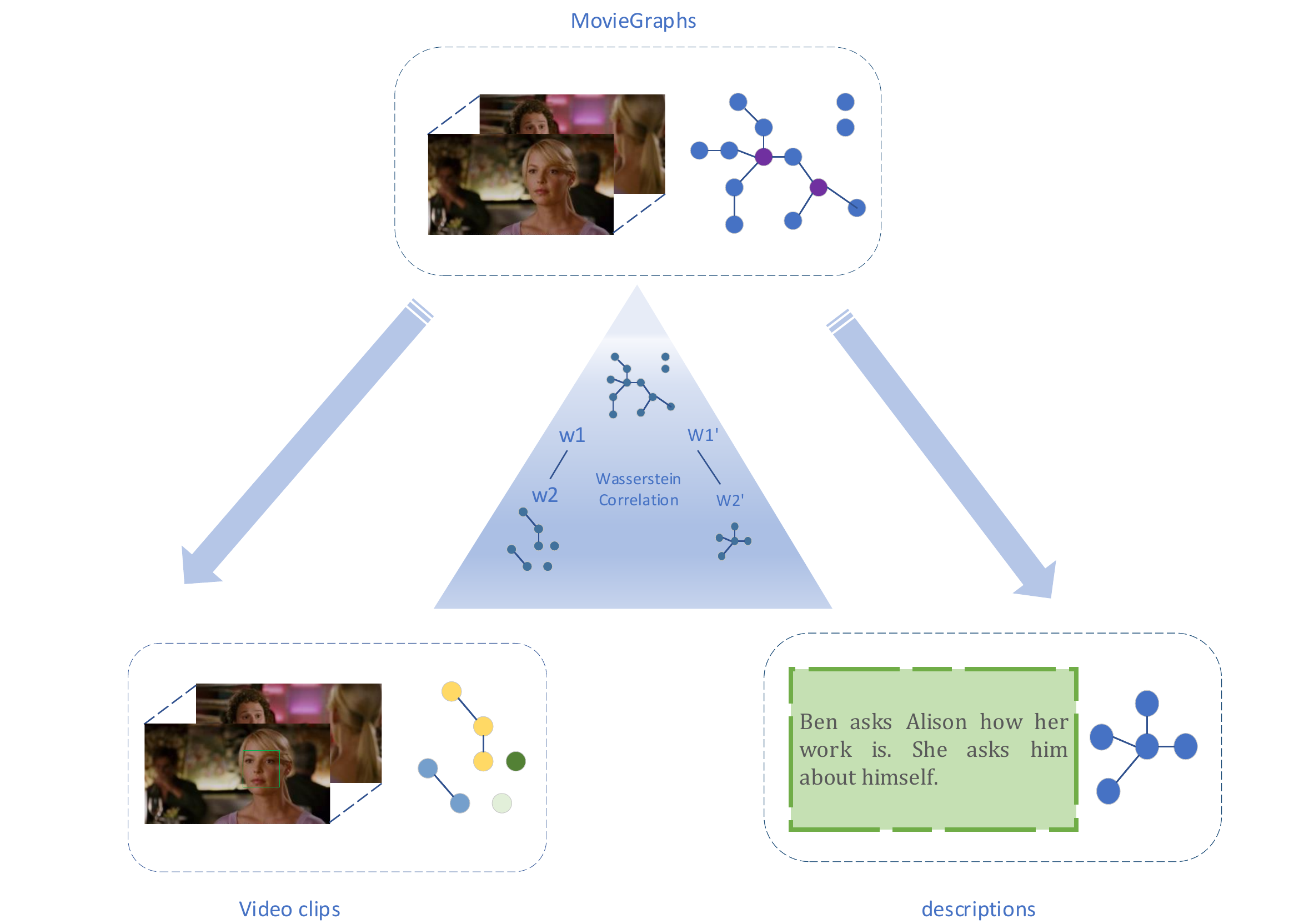}
	\caption{Our proposed GWCA is used in two retrieval tasks. It jointly encapsulates graph signal filtering and Wasserstein metric learning into a unified mathematic model and W1 and W2 are learned in this process.  Section~\ref{overview} shows more details.}
\end{figure*}\label{fig-frame}
\section{Overview}\label{overview}
In our task, we need to retrieval video clips and their descriptions using manually annotated graphs in~\cite{vicol2018moviegraphs} as queries. To better analyze the correspondence between annotated graphs, descriptions and video clips, we transform them into graph structured data and the task is converted into the problem of graph searching. For each pair of samples, we let $\X_1\in\mbR^{n_1\times d_1}, \X_2\in\mbR^{n_2\times d_2}$ denote the graph of samples in each pair respectively. To represent the annotated graph and constructed ourselves, following features are taken into consideration : 1) word embeddings for the annotated graph and the description; 2)features extracted by different neural networks for the video clip. The detail of the graph construction can be found in Section~\ref{sec5}. In order to analyze graph correlation of different magnitudes features, we project them into the same space and maximize the correlation between projections. We minimize the Wasserstein distance, that is, to learn weight parameters with regard of graphs. During training, we perform the metric training with pairwise samples and get weight parameters. In the process of testing, we search the most similar clip for the query over all the other clips with learned information.
\section{Graph Correlation Analysis}
Given a pair of (heterogeneous) graphs, e.g., annotation graph versus description graph, we denote them as $\mcG_1=(\mcV_1,\A_1,\X_1)$ and $\mcG_2=(\mcV_2,\A_2,\X_2)$, where $\mcV_1, \mcV_2$ are the node sets with the node numbers $|\mcV_1|=n_1$ and $|\mcV_2|=n_2$. The adjacent matrices $\A_1\in\mbR^{n_1\times n_1},\A_2\in\mbR^{n_2\times n_2}$ record connections of edges, graph signals $\X_1\in\mbR^{n_1\times d_1}, \X_2\in\mbR^{n_2\times d_2}$ describe attributes of all nodes, where each row corresponds to the signal vector of one node therein and $d_1, d_2$ are the dimensions of signals. Our ultimate aim is to measure the distance of these two graphs $\mcG_1,\mcG_2$. Formally, we define the distance metric learning on these two graphs as
\begin{align}
\mcD(\mcG_1,\mcG_2) = \mcM(\mcF(\mcG_1),\mcF(\mcG_2)),
\end{align}
where $\mcF(\cdot)$ is a function of graph signal processing, $\mcM(\cdot)$ is a distance metric function between two graphs.

\subsection{Graph Filtering versus Graph Metric}

Below we detailedly introduce the graph signal filtering function $\mcF$ and the graph metric learning function $\mcM$, and derive the consistency of their learning process that the metric learning could be viewed as signal filtering and vice versa.

\subsubsection*{Graph Signal Filtering}

In spectral graph theory, one main operator is the graph Laplacian operator, defined as $\L = \D-\A$, where $\D\in\mbR^{n\times n}$ is the diagonal degree matrix with $D_{ii}=\sum_{j}A_{ij}$. The popular option is to normalize graph Laplacian, i.e.,
\begin{align}
\L^{\text{norm}} = \D^{-\frac{1}{2}}\L\D^{-\frac{1}{2}}= \I-\D^{-\frac{1}{2}}\A\D^{-\frac{1}{2}},
\end{align}
where each edge $A_{ij}$ is multiplied by a factor $\frac{1}{\sqrt{D_{ii}D_{jj}}}$, and $\I$ is an identity matrix. Unless otherwise specified, below we use the normalized version. Due to the symmetric and positive definite (SPD) property, the graph Laplacian $\L$ is with a complete set of orthonormal eigenvectors. Formally, we can decompose the Laplacian matrix into
\begin{align}
\L=\U\Lambda\U\tp,
\end{align}
where $\Lambda= diag([\lambda_1,\lambda_2,\cdots,\lambda_n])$ with the spectrum $\lambda_i \geq 0$. In analogy to the classic Fourier transform, the graph Fourier transform and its inverse transform are defined as~\cite{shuman2013emerging}
\begin{align}
\hbx = \U\tp\x, \quad \x = \U\hbx,
\end{align}
where $\x\in\mbR^{n}$ is a graph signal of spatial domain, and $\hbx$ is the corresponding frequency signal.

Let $\mcF(\cdot)$ denote the filter function on graph $\mcG$, we can define the frequency response on an input signal $\x$ as
$\whz(\lambda_l)=\whx(\lambda_l)\widehat{\mcF}(\lambda_l)$, and the inverse graph Fourier transform~\cite{shuman2013emerging} as $
z(i) = \sum_{l=1}^{N} \whx(\lambda_l)\widehat{\mcF}(\lambda_l)u_l(i)$,
where $\whz(\lambda_l), \whx(\lambda_l), \widehat{\mcF}(\lambda_l)$ are the Fourier coefficients w.r.t the spectrum $\lambda_l$.
In matrix form, the filtering process can be rewritten as

\begin{equation}
\z = \widehat{\mcF}(\L)\x =\U diag[\widehat{\mcF}(\lambda_1),\cdots, \widehat{\mcF}(\lambda_n)]\U\tp\x.\label{eqn:5}
\end{equation}

Given the input signal $\x$ and the output response $\z$, our aim is to learn the filter function $\widehat{\mcF}(\cdot)$ in frequency domain, which suffers high-burden eigenvalue decomposition. To bypass it, we use a low order polynomial to approximate $\widehat{\mcF}(\cdot)$, formally, $\widehat{\mcF}(\lambda_l) = \sum_{k=0}^{K-1} \theta_{k}\lambda_l^k$,
where $\theta=[\theta_0, \theta_1, \cdots, \theta_{K-1}]\tp\in\mbR^K$ is a vector of parameters w.r.t the polynomial coefficients, and $K$ is the order number. By plug it into Eqn.~(\ref{eqn:5}), we can have
\begin{equation}
	\begin{aligned}
	\z &= \U diag[ \sum_{k=0}^{K-1} \theta_{k}\lambda_1^k,\cdots, \sum_{k=0}^{K-1} \theta_{k}\lambda_n^k]\U\tp\x \nonumber\\
	&=
	\sum_{k=0}^{K-1} \theta_{k}\U diag[ \lambda_1^k,\cdots,  \lambda_n^k]\U\tp\x
	= \sum_{k=0}^{K-1}\theta_k\L^k\x.\label{eqn:9}
	\end{aligned}
\end{equation}

Further, we may extend it to multi-dimensional signals $\X$, each of which is with different parameter, formally,
\begin{align}
\z = \sum_{k=0}^{K-1}&\sum_{j=1}^{d}\Theta_{kj}\L^k\X_{\ast j}= \sum_{k=0}^{K-1}\L^k\X\w^{(k)},\\
\st,& \quad\w^{(k)}=[\Theta_{k1},\Theta_{k2},\cdots,\Theta_{kd}]\tp,
\end{align}
where $\X_{\ast j}$ takes the $j$-th column of the matrix $\X$, $\Theta$ is the parameter to be learnt, and $\w^{(k)}$ is associated to the $k$-order term of the polynomial of graph Laplacian.

\subsubsection*{Graph Wasserstein Metric Learning}

Below we derive that $\w^{(k)}$ is also the parameters to be learnt in metric learning. To simply the derivation, we consider the $k$-order case and meantime omit the superscript of $\w^{(k)}$. For a pair of graphs $\mcG_1, \mcG_2$, the filtering response in the $k$-order case may be written as
\begin{align}
\tbx_1 = \L_1^k\X_1\w_1, \quad \tbx_2 = \L_2^k\X_2\w_2,\label{eqn:tbx}
\end{align}
where $\tbx_1\in\mbR^{n_1}, \tbx_2\in\mbR^{n_2}$ are one-dimensional signal of all nodes,  and $\w_1\in\mbR^{d_1}, \w_2\in\mbR^{d_2}$ are the graph filtering parameters for the $k$-th polynomial term case.

We use the $2^{th}$ Wasserstein distance (abbreviated as $W_2$) for the output signals $\tbx_1, \tbx_2$. Note that each node only carries with one signal, and multi-channel signals could be easily extended. Formally, when all nodes of one graph is viewed a set of signals, we define second-order statistic distance as follows


\begin{align}
	\mcD \!\!= &\|\mu_1\!-\!\mu_2\|_2^2 \!+\! tr(\Sigma_1\!+\!\Sigma_2\!-\!2(\Sigma_1^{1/2}\Sigma_2\Sigma_1^{1/2})^{1/2}),\label{eqn:D}\\
	\st,&\quad \mu_1 =\frac{1}{n_1}\1_{n_1}\tp\tbx_1, \quad \mu_2 =\frac{1}{n_2}\1_{n_2}\tp\tbx_2,\label{eqn:mu}\\
	&\quad \Sigma_1=\frac{1}{n_1}(\tbx_1-\mu_1)\tp(\tbx_1-\mu_1),\label{eqn:Sigma1} \\
	&\quad \Sigma_2=\frac{1}{n_2}(\tbx_2-\mu_2)\tp(\tbx_2-\mu_2).\label{eqn:Sigma2}
\end{align}

By integrating Eqn.~(\ref{eqn:tbx}), Eqn.~(\ref{eqn:mu}), Eqn.~(\ref{eqn:Sigma1}) and Eqn.~(\ref{eqn:Sigma2}) into the distance metic in Eqn.~(\ref{eqn:D}), we can derive out the following formulas
\begin{align}
\mu_1\tp\mu_1 &= \w_1\tp\X_1\mcK_{\mu_{1}}\X_1\w_1,\qquad\qquad\qquad\quad\\
\mu_2\tp\mu_2 &= \w_2\tp\X_2\mcK_{\mu_2}\X_2\w_2,\\
\mu_1\tp\mu_2 &= \w_1\tp\X_1\mcK_{\mu_{1}\mu_2}\X_2\w_2,\\
\Sigma_1 &= \w_1\tp\X_1\mcK_{\Sigma_1}\X_1\w_1,\\
\Sigma_2 &= \w_2\tp\X_2\mcK_{\Sigma_2}\X_2\w_2,\\
(\Sigma_1\Sigma_2)^{1/2} &\geq \w_1\tp\X_1\tp\mcK_{\Sigma_1\Sigma_2}\X_2\w_2\label{eqn:S1S2}
\end{align}
where each kernel term $\mcK$ is defined as
\begin{align}
\mcK_{\mu_{1}} &\!=\! \frac{1}{n_1^2} (\L_1^k)\tp\1_{n_1}\1_{n_1}\tp\L_1^k,\\
\mcK_{\mu_2} &\!=\! \frac{1}{n_2^2} (\L_2^k)\tp\1_{n_2}\1_{n_2}\tp\L_2^k,\\
\mcK_{\mu_{1}\mu_2} &\!=\! \frac{1}{n_1n_2} (\L_1^k)\tp\1_{n_1}\1_{n_2}\tp\L_2^k,\\
\mcK_{\Sigma_1}&\!=\!\frac{1}{n_1} (\L_1^k-\frac{1}{n_1}\1\1\tp\L_1^k)\tp(\L_1^k-\frac{1}{n_1}\1\1\tp\L_1^k),\\ \mcK_{\Sigma_2}&\!=\!\frac{1}{n_2} (\L_2^k-\frac{1}{n_2}\1\1\tp\L_2^k)\tp(\L_2^k-\frac{1}{n_2}\1\1\tp\L_2^k),\\
\mcK_{\Sigma_1\Sigma_2}&\!=\!\frac{1}{\sqrt{n_1n_2}} (\L_1^k\!-\!\frac{1}{n_1}\1\1\L_1^k)\tp(\L_2^k\!-\!\frac{1}{n_2}\1\1\L_2^k).\label{eqn:ks1s1}
\end{align}

In the above formulas, we can easily derive them except Eqn.~(\ref{eqn:S1S2}). Next we give the derivation process of  $(\Sigma_1\Sigma_2)^{1/2}$ . We denote $\tbx_1'=\tbx_1-\mu_1$ and $\tbx_2'=\tbx_2-\mu_2$, and suppose the same dimensions (i.e., $n_1=n_2$)\footnote{We can pad zero values to one of them to produce the same dimensions for them.}, and then can have
\begin{align}\label{sigma}
(\Sigma_1\Sigma_2)^{1/2} & = (\frac{1}{n_1n_2}(\tbx_1')\tp\tbx_1'(\tbx_2')\tp\tbx_2')^{1/2}\\
&\geq \frac{1}{\sqrt{n_1n_2}}(\tbx_1')\tp\tbx_2',
\end{align}
where this inequation employs Cauchy inequality: $\sum_{i=1}^n a_i^2\sum_{i=1}^n b_i^2 \geq \left(\sum_{i=1}^n a_ib_i\right)^2$. Next we plug Eqn.~(\ref{eqn:tbx}) and Eqn.~(\ref{eqn:mu}) into the above equation and define the kernel term in Eqn.~(\ref{eqn:ks1s1}). After a series of derivation, we can reach the final Eqn.~(\ref{eqn:S1S2}).

Now we can obtain the upper bound of Wasserstein distance metric, i.e.,
\begin{align}
\mcD & \leq \w_1\tp\X_1\tp(\mcK_{\mu_{1}}+\mcK_{\Sigma_{1}})\X_1\w_1 \nonumber\\
&\quad + \w_2\tp\X_2\tp(\mcK_{\mu_{2}}+\mcK_{\Sigma_{2}})\X_2\w_2\nonumber\\
&\quad - 2\w_1\tp\X_1\tp(\mcK_{\mu_{1}\mu_2}+\mcK_{\Sigma_{1}\Sigma_{2}})\X_2\w_2.\label{eqn:D2}
\end{align}
The above bound is obviously the metric learning in Wasserstein space if we extend $\w_1, \w_2$ to multi-channel responses. Therefore, the Wasserstein metric learning is consistent with graph signal filtering. In other words, the aim of metric learning is to learn graph filters, and vice verse.

\subsubsection*{Wasserstein Correlation Analysis}

Given $M$ pairs of matching graphs, $\{(\mcG_1^{(m)},\mcG_2^{(m)})\}|_{m=1}^M$, we expect to learn the projection to make their as closer as possible, formally,

\begin{equation}\label{D_mat}
\arg\min_{\w_1,\w_2}\quad\sum_{m=1}^M \mcD (\mcG_1^{(m)},\mcG_2^{(m)}).
\end{equation}

We replace $\mcD$ with Eqn.~(\ref{eqn:D2}), and then the objective function can be rewritten as
\begin{equation}
\arg\min_{\w_1,\w_2}\quad \w_1\tp\mcC_1\w_1 + \w_2\tp\mcC_2\w_2 - 2\w_1\tp\mcC_{12}\w_2,
\end{equation}
where
\begin{align}
\quad \mcC_{1}&=\sum_{m=1}^M (\X_1^{(m)})\tp(\mcK_{\mu_{1}}+\mcK_{\Sigma_{1}})\X_1^{(m)},\\
\mcC_{2}&=\sum_{m=1}^M (\X_2^{(m)})\tp(\mcK_{\mu_{2}}+\mcK_{\Sigma_{2}})\X_2^{(m)},\\
\mcC_{12}&=\sum_{m=1}^M (\X_1^{(m)})\tp(\mcK_{\mu_{1}\mu_{2}}+\mcK_{\Sigma_{1}\Sigma_{2}})\X_2^{(m)}.
\end{align}

An elegant alternative of the solution is to maximize the following objective function
\begin{equation}
\arg\max_{\w_1,\w_2} \frac{\w_1\tp\mcC_{12}\w_2}{\sqrt{\w_1\tp\mcC_{1}\w_1}\sqrt{\w_2\tp\mcC_{2}\w_2}}.
\end{equation}
which finally falls into the category of canonical correlation analysis. Hence, this maximum optimization has a closed-form solution, which can be derived as the eigenvalue decomposition from
\begin{align}
\mcC_1^{-1}\mcC_{12}\mcC_{2}^{-1}\mcC_{21}\w_1=\rho^2\w_1\\
\mcC_2^{-1}\mcC_{21}\mcC_{1}^{-1}\mcC_{12}\w_2=\rho^2\w_2
\end{align}
where $\w_1, \w_2$ are eigenvectors and $\rho$ is the correlation coefficient.

Consequently, those eigenvectors with large correlation coefficients may be chosen as multi-channel projection functions. Further, with the change of the order $k$, we can learn the corresponding filtering functions also metrics.

\section{Graph Generation}\label{sec5}
In this section, we introduce how we generate graphs on the MovieGraphs dataset. As structural difference exists between videos and descriptions of movies, different graphs are constructed accordingly.

\subsection{Graph construction on videos}
Formally, a graph can be denoted as $\mathcal{G} =(\mathcal{V}, \mathcal{E})$, where $\mathcal{V}$ and $\mathcal{E}$ are the sets of nodes and edges, respectively. Following the configuration of the Moviegraphs dataset, four types of nodes, which correspond to the nodes in the manually annotated graph of the dataset, are taken into account for video clips denoted as $M$. Nodes of character and attribute are denoted with the notations $v^{ch}$ and $v^{att}$ respectively. Another two independent nodes named scene and situation are specifically denoted as  $v^{sc}$ and $v^{si}$. Below, we introduce how we learn embeddings of these nodes, and set up connections between them.

\textbf{Scene and situation.} Scene and situation provide the context of the video. For each video clip, the Resnet is used to extract features from those frames, and features of every ten frames are averaged as the representation of scene and situation denoted as $\x_{sc}\in \mathbb{R}^{2048}$ and $\x_{si} \in \mathbb{R}^{2048}$, which is similar with the previous work~\cite{vicol2018moviegraphs}.

\textbf{Character.} In the graph retrieval video task, in order to obtain the features of different types of nodes, e.g. facial expression and age, we first perform face detection on each frame~\cite{timmurphy.org}, and then construct multiple clusters where each cluster is formed by those faces belonging to the same person. Moreover, we assign each cluster with one name according to the actor list in IMDB by comparing the features between the cluster and the actor picture (also provided in IMDB). Specifically, in the process of constructing face clusters, facial features are first extracted, and accordingly the Euclidean distances are calculated between faces for comparison. Also, a threshold is set to determine whether they are the same person. For each face cluster not aligned with an actor name, we randomly choose an unassigned name in the actor list for it.

\textbf{Attribute.} For each face cluster, its attribute node include age, gender, emotion, etc. Each attribute node is represented by the extracted feature of the predicted attribute value~\cite{pennington2014glove} (e.g. "male" for gender), which is formally denoted as $\x_{att} \in \mathbb{R}^{300}$. 
These nodes form a graph where nodes with similar embeddings are connected.
\subsection{Graph construction on Descriptions}
In moviegraphs dataset, each video clip has a natural language description. To construct one graph for each description of the movie clip, after spliting the sentence and removing stopwords and notations, we statistics the total words while keep previous order to obtain a small corporus. Here we regard each word as graph node $v^{des}$, and each node in the textual graph has the representation of a fixed length by using GloVe embeddings{~\cite{pennington2014glove}}. Moreover, the intense of the edge between nodes is defined as the similarity between their embeddings.

\section{Experiments}
We conduct experiments on the MovieGraphs dataset with our proposed GWCA. The performance of GWCA is also compared with the results of those retrieval tasks in MovieGraphs~\cite{vicol2018moviegraphs}. Moreover, we conduct an ablation study to discuss the influence of different distance metrics and different orders of receptive fields.
\subsection{Dataset and settings}
MovieGraphs dataset consists of 51 movies with annotated textual description and graphs. Each movie is split into multiple rough scenes and then manually refined. As a result, the dataset contains 7637 clips in total and each clip has an annotated description and graph. There are 35 words on average in the description, and the average number of nodes per graph is also about 35. In the experiment, following the protocol in~\cite{vicol2018moviegraphs}, the dataset is split into 5050 clips for training, 1060 clips for validation and 1527 clips for testing.

Two retrieval tasks are evaluated to test the performance: (1) descriptions retrieval using annotated graphs as queries, and (2) video clips retrieval using annotated graphs as queries. In the test stage, for each query graph, we search all the descriptions/video clips to find the most similar one. Following the previous work~\cite{vicol2018moviegraphs}, we use "Recall" as the evaluation metric, and calculate the Recall@1(R@1), Recall@5(R@5) and Recall10(R@10) to explore the effectiveness of our GWCA. There R@K stands for the fraction of correct predicted results in the top K predictions.

\subsection{The Comparison Results}
\begin{figure*}[!t]
	\centering
	\includegraphics[width=0.9\linewidth]{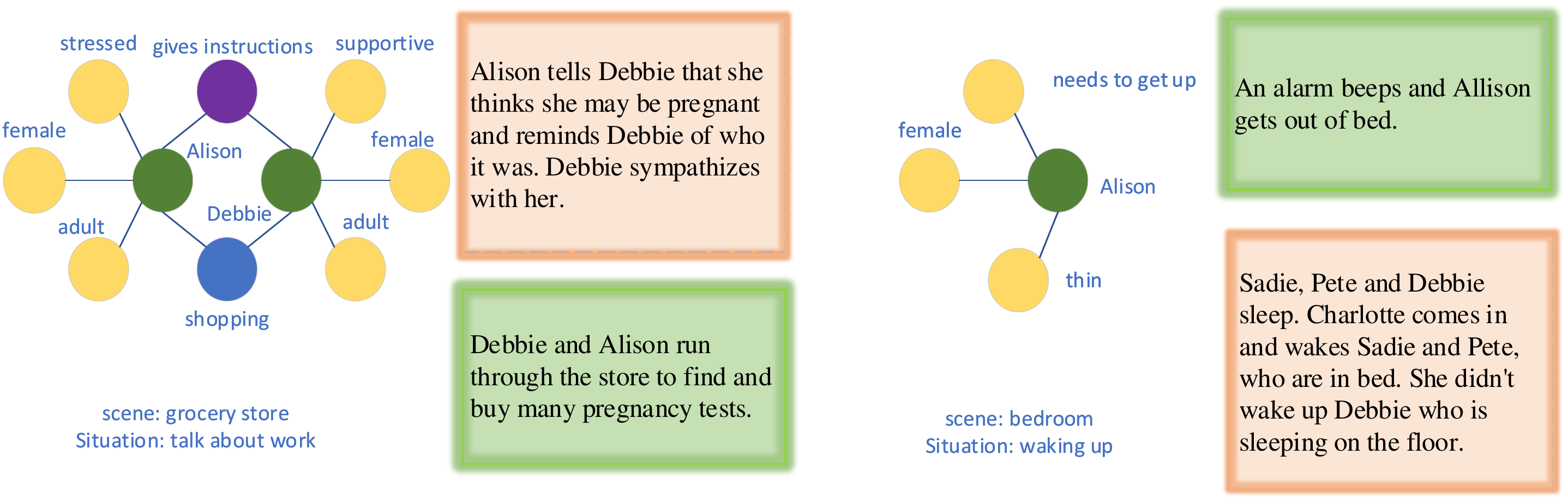}
	\caption{The results for retrieved descriptions with graphs. We show the top-2 retrieved clips. The sub-graphs indicate the query graphs. The green boxes indicate the ground-truth and the red boxes indicate the quite similar one.}
	\label{fig:2}
\end{figure*}

We compare our proposed GWCA with those state-of-the-art methods, and the results are shown in Table~\ref{tab:1}. \\
\begin{table}
	\caption{The comparison results of two different retrieval tasks.}
	\centering
    \scalebox{1.1}{
	\begin{tabular}{|c|c|c|c|c|c|c|c|}
		\hline
		\multirow{2}{*}{Method} &
		\multicolumn{3}{|c|}{\textbf{Description}} &
		\multirow{2}{*}{Method} &
		\multicolumn{3}{|c|}{\textbf{Video}} \\
		\cline{2-4}
		\cline{6-8}
		& R@1 & R@5 & R@10 &  & R@1 & R@5 & R@10\\
		\hline
		GloVe,idf$\cdot$max-sum & 61.3 & 81.6 & 86.9 & sc & 1.1 & 4.3 & 7.7 \\
		\hline
		GloVe,max-sum & 62.1 & 81.3 & 87.2 & sc,si & 1.0 & 5.4 & 8.7\\
		\hline
		TF$\cdot$IDF & 61.6 & 83.8 & 89.7 &  sc,si,a & 2.2 & 9.4 & 15.5\\
		\hline
		GWCA & $\mathbf{67.6}$ & $\mathbf{87.8}$ & $\mathbf{91.9}$ &
		sc,si,a(ours) & $\mathbf{2.4}$ & $\mathbf{10.1}$ & $\mathbf{16.2}$\\
		\hline
	\end{tabular}}
	\label{tab:1}
\end{table}
\textbf{Description Retrieval using graphs as queries.} The results of description retrieval with query graphs are shown in Table~\ref{tab:1}, from the first row to third row in the second column. For the compared methods, GloVe means that the GloVe word embedding is employed; max-sum and idf$\cdot$max-sum are pooling strategies with word embedding. Specifically, idf$\cdot$max-sum weights words with rarity. The previous method~\cite{vicol2018moviegraphs} finds the best matching word in description for each word in the manually annotated graph, and sum up them to compute the similarity. Then this processed score is fed into the loss function. According to Table~\ref{tab:1}, GloVe with the pooling strategy of max-sum achieves limited performance gain comparing with GloVe. TF$\cdot$IDF, which uses an identity sparse matrix to initialize features, performs better than GloVe. Among these compared methods, our GWCA shows the best performance, where the score of Recall@1 is about 5.5$\%$ higher than GloVe with max-sum pooling. Besides, for those words with similar meaning/embedding which are sometimes confusing, our GWCA is still effective enough to compute the correlation and fulfill the retrieval task well. This observation demonstrates that GWCA successfully formulates graph signal encoding together with graph distance metric learning into a unified model. Besides, we show some retrieval examples in Fig.~\ref{fig:2}.

\begin{figure*}[!t]
	\centering
	\includegraphics[width=0.9\linewidth]{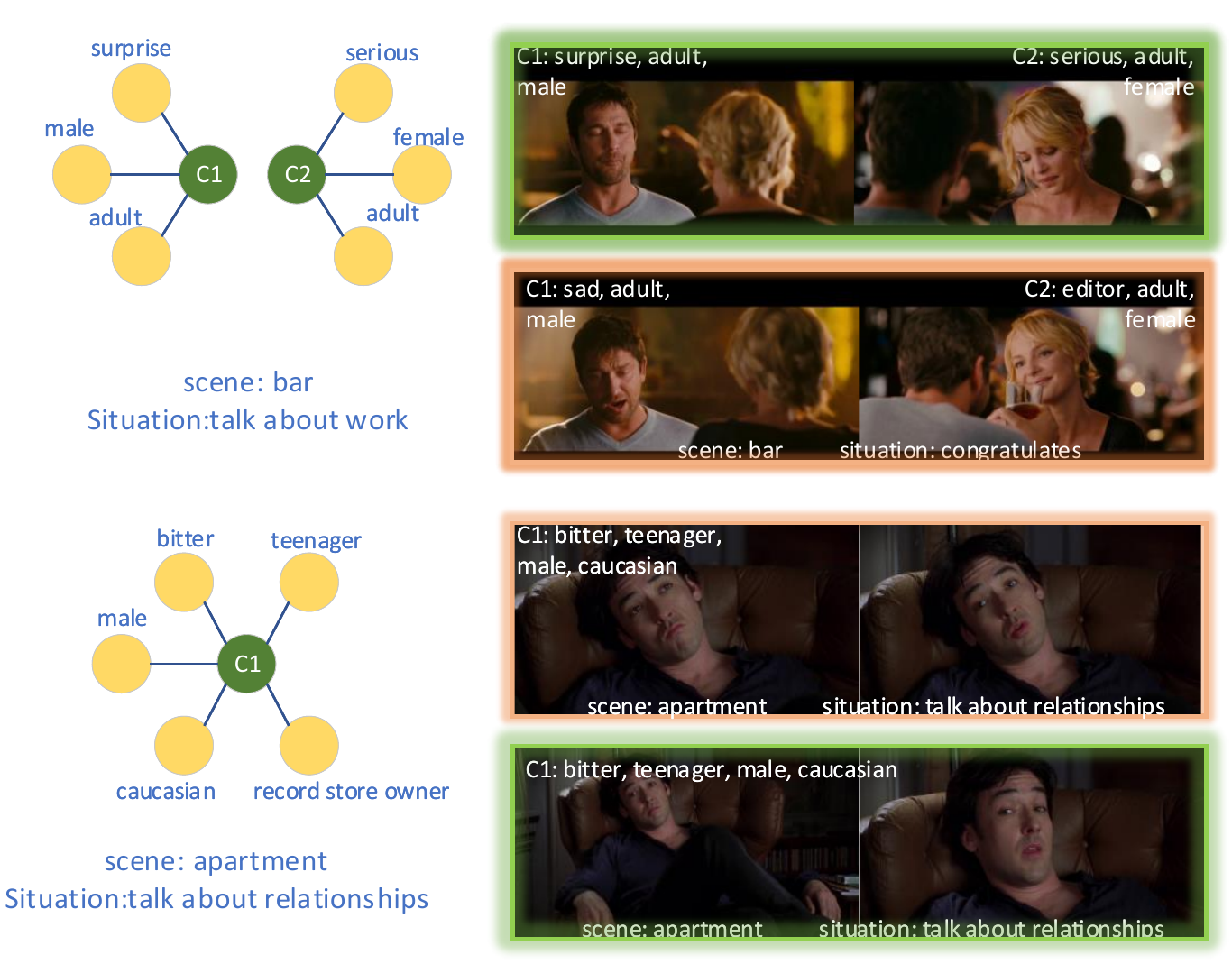}
	\caption{The results for retrieved video clips with graphs. We show the top-2 retrieved clips. The red boxes indicate results that are quite similar in meaning to the query, and the green boxes indicate ground-truth.  }
	\label{fig:3}
\end{figure*}

\textbf{Video Retrieval using graphs as queries.} This experiment aims to measure the performance of our method to retrieve videos based on the given annotated graphs. The result is shown in the third and fourth columns in Table~\ref{tab:1}. 
The characters 'si', 'sc' and 'a' indicate that we start with the scene, situation, attributes, and characters as part of graphs, while their corresponding methods all compute the cosine similarity to measure the distances between nodes. The reported result of our GWCA employs all the four kinds of nodes, and achieves the best performance. According to the shown result, we have the following observations: (1) the four different nodes, i.e. situation, scene, attribute and characters, all contribute to the video retrieval; (2) compared with other methods, our GWCA is advantageous in understanding the graph structure as it jointly encapsulates graph signal filtering and Wasserstein metric learning into a unified mathematic model helps to enhance the node representation ability. 
Some examples of the retrieved videos are visualized in Fig.~\ref{fig:3} for the intuitive impression of our GWCA. 

\subsection{Ablation Study}
\begin{figure}
	\begin{minipage}[t]{0.5\linewidth}
		\centering
		\includegraphics[width=2.2in]{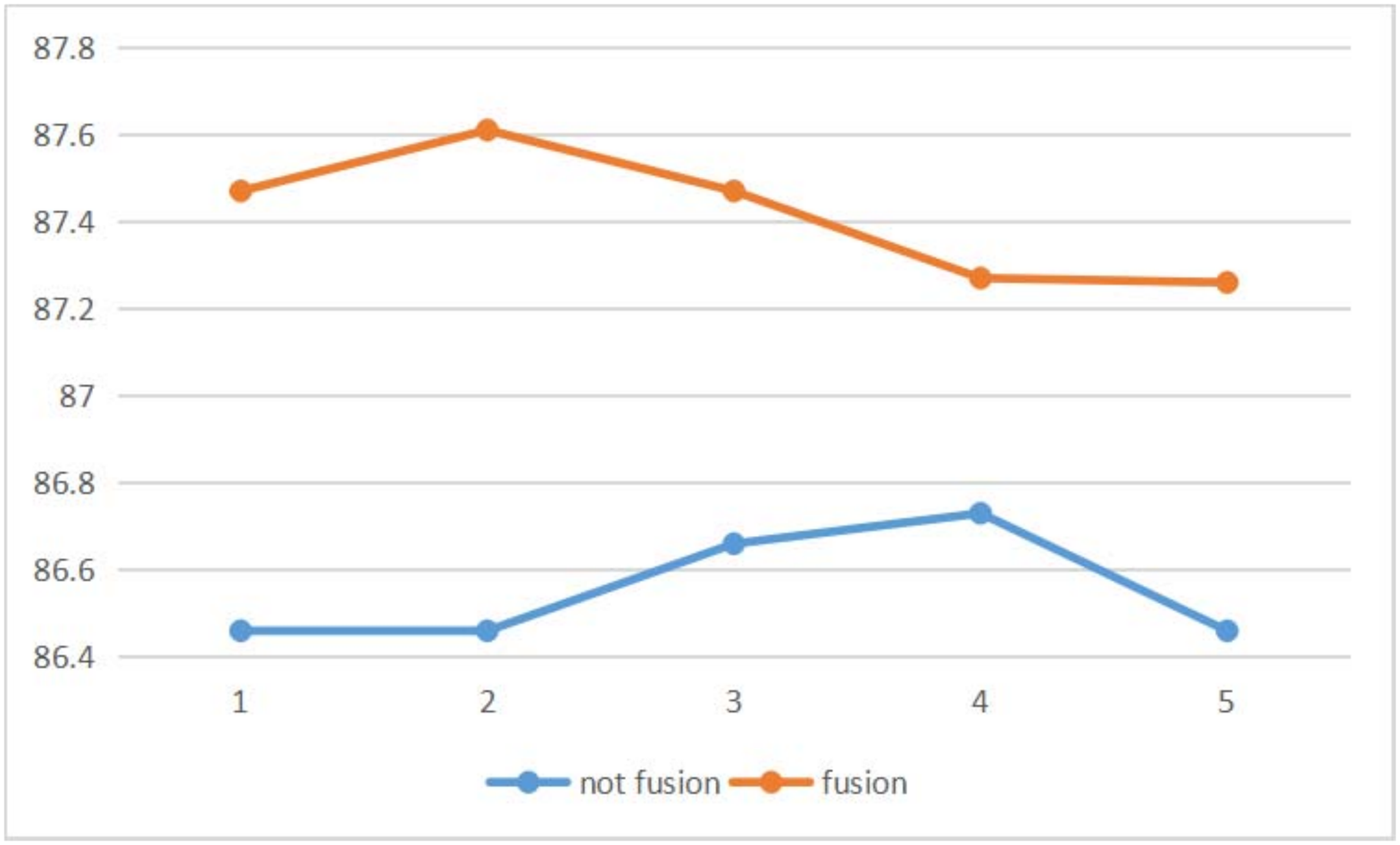}
		\caption{Test Recall@5 with different \protect\\orders, using GWCA in the description \protect\\retrieval task.}
		\label{fig:4}
	\end{minipage}%
	\begin{minipage}[t]{0.5\linewidth}
		\centering
		\includegraphics[width=2.2in]{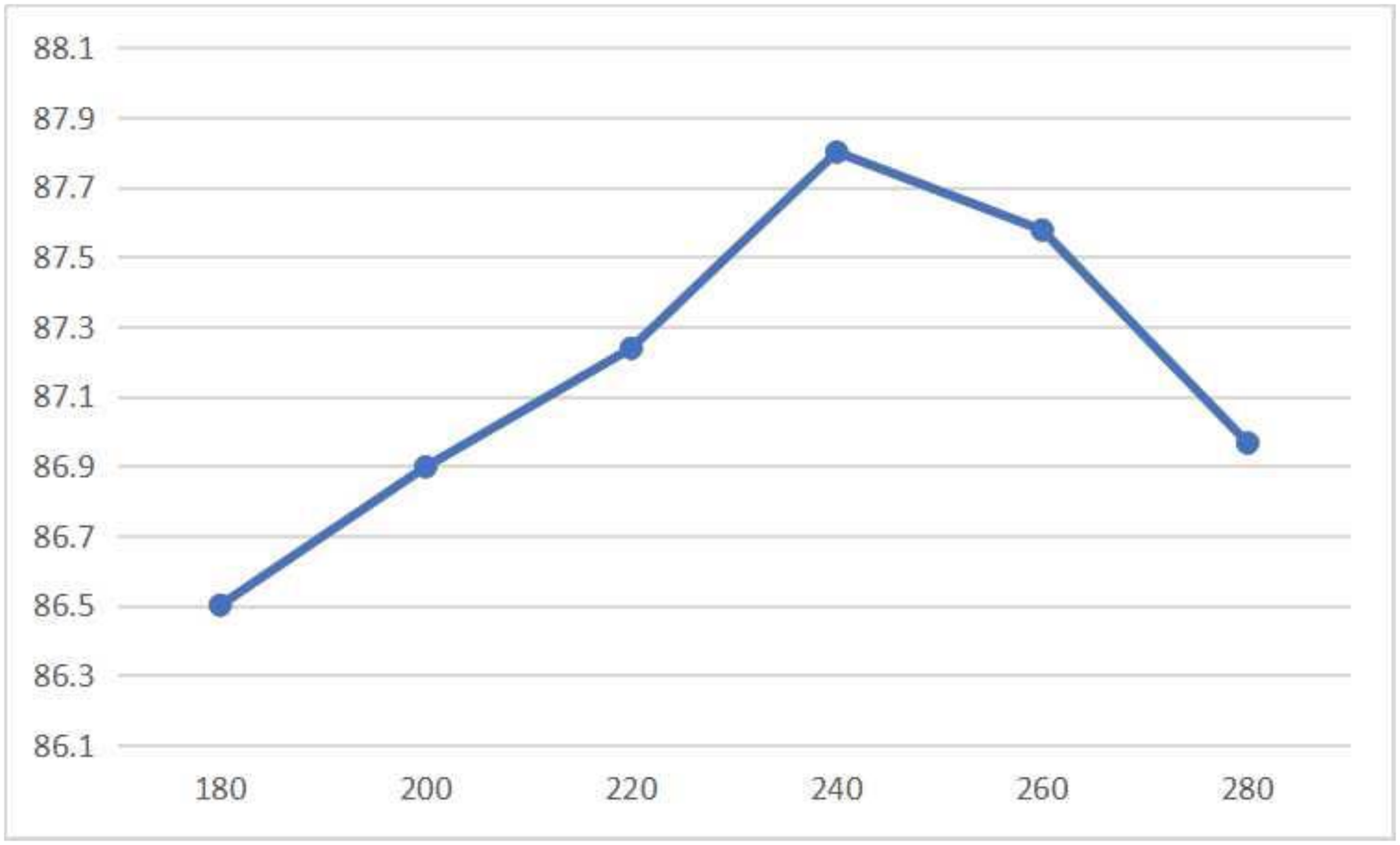}
		\caption{Test Recall@5 with different dimensions, using GWCA in the description retrieval task.}
		\label{fig:5}
	\end{minipage}
\end{figure}

In this section, we dissect our algorithm by conducting ablation analysis. Specifically, we evaluate how the modules, i.e. the graph Wasserstein metric, the order of receptive fields, and the dimension of the features of nodes, promote the retrieval. For this purpose, we conduct the following additional experiments:
\begin{enumerate}
	\item[(1)] Comparing the performance between Graph Wasserstein Metric and cosine distance with different algorithms, e.g. PCA, CCA, and also the original feature without learning. The result is shown in Table~\ref{tab:2}.
	\item[(2)] Comparing the performance of GWCA under different values of the order of receptive field $k$. Please see the result in Fig.~\ref{fig:4}. Fusion means the features of those orders that are lower than $k$ are fused together as the feature.
	\item[(3)] Comparing the performance of GWCA under different dimensions of node features. The result is shown in Fig.~\ref{fig:5}.
\end{enumerate}

According to the results, we have the following observations:

\begin{enumerate}
	\item[(1)]  Graph Wasserstein Metric effectively promotes the performance in distance measurement between graph. Specifically, for both PCA and CCA, higher performances are achieved with Graph Wasserstein Metric than cosine similarity.
	\item[(2)]  The order of receptive field k also influences the performance. We focus on the situation of fusion as it achieves higher performance in Fig.~\ref{fig:4}. As it is shown, we can see that the best performance is achieved when k equals 2, otherwise the performance drops.
	\item[(3)] The dimension of node feature is also an important factor influencing the retrieval performance. According to Fig.~\ref{fig:5}, the performance varies with different dimensions, and the best performance is achieved when the dimension is set to 240.
\end{enumerate}

\begin{table}[!htbp]
    \centering
    \caption{The comparison results between cosine distance and Graph Wasserstein Metric using different algorithms.}
	\setlength{\tabcolsep}{6mm}{
	\begin{tabular}{|c|c|c|c|c|}	
		\hline
		\multicolumn{2}{|c|}{ \multirow{1}*{Method} }&R@1&R@5&R@10\\
		\cline{2-5}
		\hline
		Ori Feature&cos&4.07&12.4&19.6\\
		\cline{1-5}
		\multirow{2}*{PCA}&cos&35.9&55.5&64.8\\
		\cline{2-5}
		&w-2&62.1&77.7&86.7\\
		\hline
		\multirow{2}*{CCA}&cos&61.9&78.3&85.3\\
		\cline{2-5}
		&w-2&66.4&82.7&88.3\\
		\hline
		\multicolumn{2}{|c|}{GWCA} & $\textbf{67.6}$ & $\textbf{87.8}$ & $\textbf{91.9}$ \\
		\hline
	\end{tabular}}
    \label{tab:2}
\end{table}

\section{Conclusion}
In this paper, a Graph Wasserstein Correlation Analysis (GWCA) method was proposed to deal with the comparisons of pairwise movie graphs and show the effectiveness. We relabel some content ourselves after downloading the existing data, and then use GWCA to formulate graph signal encoding together with graph distance metric learning on this dataset. In this way, graph signal filtering and Wasserstein metric learning are jointly encapsulated into a unified model, which efficiently preserves the proximity of the nodes of graph and empowering the learned node representations. Extensive experiments and our visualizations analyze our method and we believe that our contribution can be applied to many domains.
\section*{Acknowledgments}
This work was supported by the National Natural Science Foundation of China (Grants Nos.  61906094, 61972204), the Natural Science Foundation of Jiangsu Province (Grant Nos. BK20190019, BK20190452), and the fundamental research funds for the central universities (No. 30919011232).

%
%
%
%
\bibliographystyle{splncs04}
\bibliography{reference}

\end{document}